\pdfoutput=1
\PassOptionsToPackage{table,xcdraw}{xcolor}
\documentclass[10pt,twocolumn,letterpaper,table,xcdraw]{article}

\usepackage{iccv}
\usepackage{times}
\usepackage{epsfig}
\usepackage{graphicx}
\usepackage{amsmath}
\usepackage{amssymb}
\usepackage{fancybox,framed}
\usepackage{tabularx}
\usepackage{array}
\usepackage{booktabs}

\usepackage{authblk}
\usepackage{marvosym}
\usepackage{marginnote}

\usepackage[accsupp]{axessibility}  

\usepackage{verbatim}
\usepackage{multirow}
\usepackage[table,xcdraw]{xcolor}

\usepackage{pifont}
\newcommand{\cmark}{\ding{51}}%
\newcommand{\xmark}{\ding{55}}%

\usepackage{booktabs}

\usepackage{wasysym}
\usepackage{bm}
\usepackage{soul}
\setstcolor{orange}

\newcommand{\deeppac}{{\sc DeepPAC}\xspace}
\newcommand{\trajpac}{{\sc TrajPAC}\xspace}
\DeclareMathOperator\dif{d\!}

\usepackage[pagebackref=true,breaklinks=true,letterpaper=true,colorlinks,bookmarks=false]{hyperref}

\usepackage[capitalise]{cleveref} 
\Crefname{section}{Sect.}{Sects.}
\Crefname{figure}{Fig.}{Figs.}
\Crefname{table}{Tab.}{Tabs.}

\iccvfinalcopy 

\def\iccvPaperID{10836} 

\ificcvfinal\fi

\newtheorem{thm}{Definition}

\newcommand{\gaojie}[1]{{\color{blue}{}#1}}
\newcommand{\nxu}[1]{{\color{orange}{}#1}}
\newcommand{\p}{\mathrm{p}}
\newcommand{\f}{\mathrm{f}}
\newcommand{\ADE}{\mathrm{ADE}}

\begin{document}

\title{\trajpac: Towards Robustness Verification of Pedestrian Trajectory Prediction Models}


\author[1,2]{Liang Zhang}
\author[1]{Nathaniel Xu}
\author[1]{Pengfei Yang}
\author[1\Letter]{Gaojie Jin}
\author[3]{Cheng-Chao Huang}
\author[1,2\Letter]{Lijun Zhang}
\affil[1]{State Key Laboratory of Computer Science, Institute of Software, CAS, Beijing, China}
\affil[2]{University of Chinese Academy of Sciences, Beijing, China}
\affil[3]{Nanjing Institute of Software
Technology, ISCAS, Nanjing, China
\vspace{1mm}
\authorcr

\tt\small \{zhangliang,gaojie,zhanglj\}@ios.ac.cn\authorcr
\textsuperscript{\Letter} Corresponding Author
}



\maketitle
\ificcvfinal\thispagestyle{empty}\fi

\begin{abstract}

Robust pedestrian trajectory forecasting is crucial to developing safe autonomous vehicles. 
Although previous works have studied adversarial robustness in the context of trajectory forecasting, some significant issues remain unaddressed. In this work, we try to tackle these crucial problems.
Firstly, the previous definitions of robustness in trajectory prediction are ambiguous. 
We thus provide formal definitions for two kinds of robustness, namely label robustness and pure robustness. 
Secondly, as previous works fail to consider robustness about all points in a disturbance interval, we utilise a probably approximately correct (PAC) framework for robustness verification. 
Additionally, this framework can not only identify potential counterexamples, but also provides interpretable analyses of the original methods.
Our approach is applied using a prototype tool named \trajpac. With \trajpac, we evaluate the robustness of four state-of-the-art trajectory prediction models --- Trajectron++, MemoNet, AgentFormer, and MID --- on trajectories from five scenes of the ETH/UCY dataset and scenes of the Stanford Drone Dataset. 
Using our framework, we also experimentally study various factors that could influence robustness performance.

\end{abstract}
 \vspace{-1 em}
\section{Introduction}
\label{sec:intro}

\begin{figure}[!t]
\includegraphics[width=0.46
\textwidth]{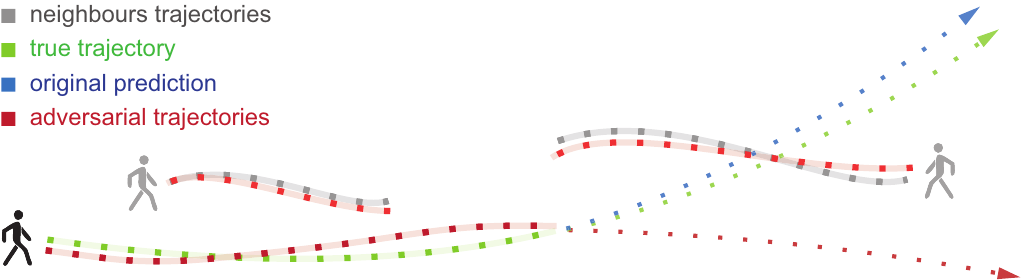}
\centering
\caption{\centering An example of adversarial attacks} 
\label{fig:path}
\vspace{-1.2em}
\end{figure}

Forecasting the movements of people based on their past states is a crucial task in both human behavior comprehension and self-driving systems~\cite{levinson2011towards}. 
This task is commonly referred to as pedestrian trajectory prediction. Although current methods~\cite{li2022graph,xu2022groupnet,bae2022non,duan2022complementary,shi2021sgcn,mohamed2020social,mangalam2020not} for predicting human trajectory have achieved remarkable results, they still face security risks due to their susceptibility to adversarial attacks. As \Cref{fig:path} shows, even a slight and hardly perceptible alteration in the previous state can lead to a significant variation in the prediction result. 

Several works~\cite{zhang2022adversarial,cao2022advdo,jiao2022semi,cao2022robust,zheng2023robustness,tan2022targeted} in the literature study the robustness of trajectory prediction models through the lens of adversarial attack and defense. However, many of these methods are directly translated from problems in image classification and still do not fully consider the specific circumstances of trajectory prediction tasks. As such, they have several overlooked shortcomings for benchmarking the robustness in forecasting problems. To this end, this work endeavors to both theoretically and experimentally analyse and mend these flaws.

The first problem is the current research \textbf{does not provide an exact and formal definition of robustness} for trajectory prediction tasks.
They emphasise that the adversarial trajectory is ``natural and feasible" \cite{zhang2022adversarial} or ``close to the nominal trajectories" \cite{cao2022advdo} but lacks a mathematical definition for what constitutes robustness (i.e., notion of robustness radius). Unlike the robustness of classification tasks, trajectory prediction is framed as a regression problem. As such, directly translating the definition of robustness from image classification to trajectory prediction is nontrivial. I.e., at what level of alignment between the prediction and ground truth can the model be deemed robust? For this reason, we provide a formal definition (\cref{subsec: definitions of robustness}) that explicitly defines the acceptable perturbation radius of historical trajectories. Our definition formally unifies the semantic definitions of robustness in previous works.

Secondly, the current research only evaluates the effectiveness of attacks by measuring the difference between the post-attack predicted path and the ground truth, \textbf{but fails to take into account the difference between the post-attack prediction and the pre-attack prediction.} It is unclear whether robustness should be measured by the difference between post-attack output and pre-attack output, or the gap between post-attack prediction and ground truth.
In order to address this issue, we present two novel definitions of robustness: label robustness, which quantifies robustness in prediction \emph{accuracy} after attacks; and pure robustness, which measures robustness in prediction \emph{stability} after attacks.

It should be noted that due to the inherent indeterminacy in human behavior, numerous stochastic prediction techniques have been introduced to capture the multi-modality of future movements. Even for unperturbed examples, the predictions of these models at identical inputs may be different. This presents a challenge to our definition of pure robustness. To address this issue, we propose to compare post-attack predictions with the empirical distribution of pre-attack predictions. The pure robustness can then be thought of as a measure of \emph{disjointness} between an adversary and the model's original forecast distribution.

Thirdly, the current literature on robustness in trajectory prediction focuses on benchmarking susceptibility to adversarial attacks, while overlooking the more rigorous problem of verification. That is to say, \textbf{current works fail to consider robustness about all points in a disturbance interval}. This is largely due to the computational infeasibility of such a procedure in continuous state spaces. To make verification more practical, we take inspiration from \deeppac~\cite{li2022towards} and probabilistically relax our definitions of robustness. In doing so, we allow efficient verification in a probably approximately correct (PAC) framework. We quantify the uncertainty associated with our method with PAC guarantees on the confidence and error rate. Moreover, our method involves learning a PAC locally linear model, which we show can be leveraged to find adversaries comparable to those found in classical attack methods like projected gradient descent~\cite{madry2017towards}.

Finally, there is a \textbf{lack of exploration into the interpretability of adversarial attacks} on trajectory prediction models. Oftentimes, perturbations added to one feature have greater influence on the output than perturbations added to other features. For example, in trajectory forecasting one might expect noise at the agent's current position to have greater impact on the output than noise added to the agent's original position. Using our PAC linear model, we aim to identify the features most sensitive to perturbation and provide an interpretable explanation for our findings. Moreover, our interpretability analysis provides a stronger understanding of what trajectory forecasting models ``see" when making future predictions.

 Our main contributions are summarised as follows:
\begin{enumerate}

\item To the best of our knowledge, we are the first to formally define robustness for trajectory prediction models, namely label robustness and pure robustness, which allows us to specify the prediction accuracy and stability of the models after attacks. (\Cref{section: Formulation})

\item We propose \trajpac, a framework for robustness verification of trajectory forecasting models. It takes inspiration from \deeppac~\cite{li2022towards} in that we regard the complex trajectory prediction model as a black box and learn a local PAC model by sampling. Due to the stochasticity in trajectory forecasting models, this generalisation is theoretically non-trivial. With the learned PAC model, we show how to conduct the analysis of robustness and interpretation for trajectory prediction models. (\Cref{sec:model})

\item  We use \trajpac to evaluate the robustness of four state-of-art trajectory forecasting models on the ETH/UCY dataset and three of them on the Stanford Drone Dataset. Our \trajpac shows good scalability on various trajectory forecasting models and different robustness properties. It is highly efficient, as the running time for model learning and verification is within seconds. Although \trajpac only provides a PAC guarantee, we claim that it is empirically sound because no counterexamples can be found by PGD~\cite{madry2017towards} on all the cases where the PAC model learned by \trajpac is robust. Also, we find that \trajpac is capable of finding adversarial examples comparable to PGD. Through an interpretation analysis, we study the potential factors that contribute to robustness. (\Cref{sec: Experiment})

\end{enumerate}

\section{Related Work}
\label{sec:related}

\textbf{Pedestrian Trajectory Prediction.}
Based on the observed paths, the goal of a human trajectory forecasting system is to estimate future positions. Early work in trajectory prediction utilised deterministic approaches such as social forces~\cite{helbing1995social,mehran2009abnormal}, Markov processes~\cite{kitani2012activity,wang2007gaussian}, and RNNs~\cite{alahi2016social,morton2016analysis,vemula2018social}. However, as human behavior is inherently unpredictable, numerous stochastic prediction methods have been proposed to model the multiple possible outcomes of future movements. Among these methods, works utilizing generation frameworks, such as \cite{dendorfer2021mg,fang2020tpnet,gupta2018social,kosaraju2019social,sadeghian2019sophie,sun2020reciprocal,zhao2019multi,amirian2019social,fernando2019gd,li2019conditional,vaswani2017attention} using GAN~\cite{goodfellow2020generative} and \cite{chen2021personalized,ivanovic2019trajectron,lee2017desire,liu2021social,salzmann2020trajectron++,tang2019multiple,yuan2021agentformer} using CAVE~\cite{sohn2015learning}, have achieved good experimental performance.
 Recently, new methods like \cite{cheng2021exploring,mangalam2020not,sun2021three} using Encoder-Decoder structures have been applied to this task because of the flexibility of these structures in encoding various contextual features. MID~\cite{gu2022stochastic} proposes a new stochastic framework with motion indeterminacy diffusion, which formulates the trajectory prediction problem as a process from an ambiguous walkable region to the desired trajectory. In contrast to parameter-based frameworks that optimize model parameters using training data, Memonet~\cite{xu2022remember} proposes a new instance-based framework based on retrospective memory, which memorizes various past trajectories and their corresponding intentions.
In this work, we choose to analyse the robustness of four distinct multi-modal prediction methods: Trajectron++~\cite{salzmann2020trajectron++}, MemoNet~\cite{xu2022remember}, AgentFormer~\cite{yuan2021agentformer} and MID~\cite{gu2022stochastic}.

\textbf{Adversarial Robustness.}
Deep learning models have been demonstrated to be susceptible to adversarial attacks \cite{carlini2019evaluating,demontis2019adversarial,carlini2017towards,xiao2018generating,yang2020patchattack,xie2017adversarial,huang2020universal,xiang2019generating,wen2020geometry,hamdi2020advpc,xiao2019meshadv,jin2022enhancing,jin2023randomized}. 
However, in the context of autonomous vehicles, there's little study on the adversarial robustness of trajectory forecasters. Several studies~\cite{zhang2022adversarial,cao2022advdo,jiao2022semi,cao2022robust,zheng2023robustness,tan2022targeted} have examined the adversarial robustness of trajectory prediction models using the lens of adversarial attack and defense, but these studies still experience essential flaws that we have detailed in \Cref{sec:intro}. Traditional verification methods~\cite{boopathy2019cnn,katz2019marabou,li2020prodeep,singh2018fast,singh2019abstract,tran2020nnv,yang2021improving,katz2017reluplex} can provide guaranteed robustness verification results, but they are unable to deal with the size of modern neural networks. Statistical methods are proposed in \cite{baluta2021scalable,baluta2019quantitative,cardelli2019statistical,mangal2019robustness,webb2018statistical,weng2019proven,wicker2020probabilistic,li2022towards} to assess the local robustness of deep neural networks with a probably approximately correct (PAC) guarantee, namely the network satisfies a probabilistic robustness property with a certain level of confidence. This type of method can better address the limitations of traditional robustness verification methods. In this work, we conduct research in this direction to investigate the robustness of trajectory prediction.

\textbf{Interpretation Analysis.}
Deep learning systems have led to significant advancements in many aspects of our lives. However, their black-box nature poses challenges for many applications. It is generally difficult to rely on a system that cannot provide explanations for its decisions. This has spurred a substantial amount of research on explainable AI methods~\cite{lam2021finding,wang2021interpretable,nauta2021neural,carter2021overinterpretation,parekh2021framework,ismail2021improving,jeyakumar2020can,shrikumar2017learning,hendricks2016generating,alvarez2018towards,huang2022safari,zhao2021baylime}, which supplement network predictions with explanations that humans can understand. However, there is currently limited research focused on providing explanations for the trajectory prediction of different methods. In our study, we train a PAC model to offer an interpretable analysis of the original model.

\section{Problem Formulation}
\label{section: Formulation}

In this section, we present the formal modeling of trajectory prediction models and the formal specification of robustness in such models.

\subsection{Trajectory Prediction}

Denote by $x^t \in \mathbb{R}^2$ the spatial coordinate of an agent at timestamp $t$, then a trajectory over $T$ timestamps is a sequence of the coordinates represented by a matrix $X\in \mathbb{R}^{2\times T}$. 
Considering the current timestamp as $t=0$, we mark the timestamps as  $t=-T_\p+1,-T_\p+2,\ldots,0$ for a past trajectory over $T_p$ timestamps. Then, let $X_0\in \mathbb{R}^{2\times T}$ be
the past trajectory of the to-be-predicted agent and $X_1,X_2,\ldots,X_N$ be those of $N$ neighbouring agents.
For $t=1,2,\ldots,T_\f$, we use $Y_\f$ to denote the ground truth of the future trajectory of the to-be-predicted agent.
The goal of trajectory prediction is to train a prediction model $g:(\mathbb{R}^{2\times T_\p})^{N+1}\to \mathbb{R}^{2\times T_\f}$, so that the predicted future trajectory 
$Y=g(X_0,X_1,\ldots,X_N)$ 
 is as close to the ground-truth $Y_\f$ as possible.

In current trajectory prediction models, stochastic prediction techniques have been introduced to capture the multi-modality of future movements, so the output of such trajectory prediction models are not deterministic, but probabilistic. In this work, due to the random mechanism widely used in trajectory prediction models, we consider the output $g(X_0,\ldots,X_N)$ of a trajectory prediction model as a probability distribution on the Borel measurable space $\mathbb{R}^{2\times T_\f}$. We write $Y \in g(X_0,\ldots,X_N)$ if $Y$ is in the support of the probability distribution $g(X_0,\ldots,X_N)$.

\subsection{Robustness of Prediction Models}
\label{subsec: definitions of robustness}

Although existing works have explored the robustness of trajectory prediction models \cite{zhang2022adversarial,cao2022advdo,jiao2022semi,cao2022robust,zheng2023robustness,tan2022targeted}, they fail to provide a formal definition of robustness. Instead, these works quantify robustness by their vulnerability to adversarial attacks. Therefore, we first provide rigorous definitions for robustness in the context of trajectory forecasting.

To describe a robustness region, we employ the $L^\infty$-norm, which is most often used in robustness verification. For an input trajectory $\hat X\in\mathbb{R}^{2\times T}$, we consider any spatial coordinate $x^t$ of the trajectory can be disturbed in the closed $L^\infty$-norm ball with the center $x^t$ and the radius $r>0$.
Then, we use $B(\hat X, r)$ to denote the set of the disturbing trajectories generated from $\hat X$, i.e.,
 $B(\hat X, r)=\{X\in\mathbb{R}^{2\times T} \mid ||X-\hat X||_\infty \le r\}$. 


Since trajectory prediction models are regression models, we cannot define local robustness as that in classification tasks, where the robustness property can be naturally given with the output scores. 
To define robustness in trajectory prediction models, we adapt the same intuition as that in global robustness~\cite{DBLP:conf/ijcai/RuanWSHKK19}, which requires that the output perturbation should be uniformly bounded.
The output perturbation can be formalised as a metric $D:\mathbb{R}^{2\times T_\f} \times \mathbb{R}^{2\times T_\f} \to [0,+\infty)$.
If we use the ground truth $Y_\f$ to measure the output perturbation, we have the following definition of label robustness:



\begin{thm}[Label Robustness]  \label{def:labelrobustness}
Let $\hat {\mathbf X}=(\hat X_0,\hat X_1,\ldots,\linebreak[0] \hat X_N)$ be the past trajectories of the to-be-predicted agent and its $N$ neighbouring agents, and
$Y_\f$ is ground truth of the future trajectories of the to-be-predicted agent.
Given a prediction model $g$, an evaluation metric $D$, a safety constant $s$, then $g$ is label-robust at $\hat {\mathbf X}$ w.r.t. the perturbation radius $r > 0$ if for any $X_i\in B(\hat X_i,r)$ ($i=0,1,\ldots,N$) and any $Y \in g(X_0,X_1,\ldots,X_N)$, we have $D(Y, Y_\f) \leq s$.
\end{thm}

Robustness in models with random mechanism is quite different, where we require that $D(Y, Y_\f) \leq s$ for any possible output trajectory $Y$.
In Def.~\ref{def:labelrobustness}, we always assume that the input $\mathbf X$ is chosen from the dataset, so that its ground truth $Y_\f$ is accessible. 
Since we measure the distance from the ground truth, a label-robust model intuitively has good performance in prediction and tolerance to adversaries.
However, label robustness has the limitation that we must have the ground truth $Y_\f$, so it is difficult to adapt it to the robustness regions where we do not know the ground truth.
For such a consideration, we define pure robustness, where distance is measured from the output of $\hat {\mathbf X}$ in the model:

\begin{thm}[Pure Robustness] \label{def:purerobustness}
Let $\hat {\mathbf X}=(\hat X_0,\hat X_1,\ldots,\linebreak[0] \hat X_N)$ be the past trajectories of the to-be-predicted agent and its $N$ neighbouring agents.
Given a prediction model $g$, an evaluation metric $D$, a safety constant $s$, then $g$ is purely robust at $\hat {\mathbf X}$ w.r.t. the perturbation radius $r > 0$ if for any $X_i\in B(\hat X_i,r)$ ($i=0,1,\ldots,N$) and any $Y \in g(X_0,X_1\ldots,X_N)$, there exists $\hat Y \in g(\hat {\mathbf X})$, s.t. $D(Y,\hat Y) \leq s$.
\end{thm}

\begin{figure}[tbp]
\includegraphics[width=0.48
\textwidth]{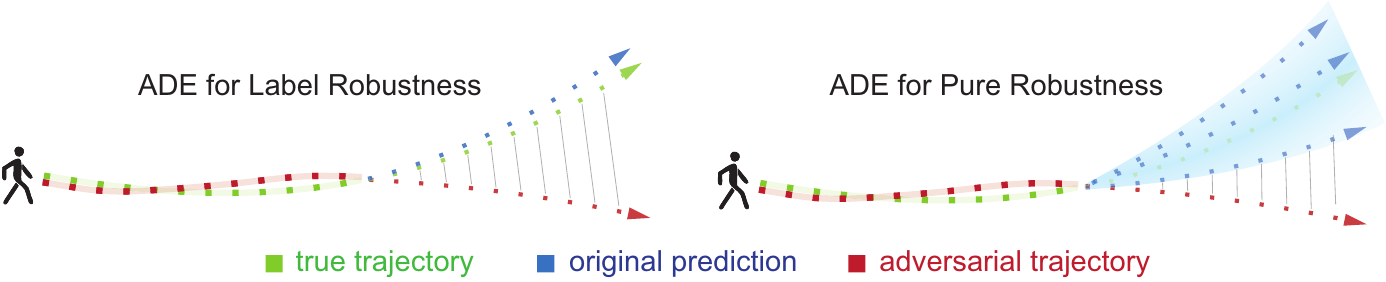}
\centering
\caption{ADE of the adversarial trajectory prediction for label robustness and pure robustness.
}
\label{fig:ADE_2}
\vspace{-1 em}
\end{figure}

We call it pure robustness since the distance is measured from the output of the model, in which situation only tolerance to adversaries is described. 
In Def.~\ref{def:purerobustness}, we make more modifications for the random mechanism, since the output $g(\mathbf X)$ is also a distribution. 
For an output trajectory $Y$, we look for a trajectory $\hat Y \in g(\hat {\mathbf X})$ such that their distance attains the minimum, and pure robustness requires that this minimum distance should be smaller than the safety constant $s$.
In \Cref{fig:ADE_2} we show the difference between label robustness and pure robustness.


To specify the definition of robustness, we still need to determine the evaluation metric $D$ to measure the difference of two trajectories.
Here we employ Average Displacement Error (ADE)~\cite{alahi2014socially,alahi2016social,gupta2018social,li2019conditional}, which refers to the mean $L^2$ distance between all coordinates of ground truth and those of
the predicted trajectory. For two trajectories $Y_1$ and $Y_2$ over timestamps $t=1,2,\ldots,T$ , we generalise ADE with $L^2$ norm to measure the distance between them:
\[
\ADE(Y_1, Y_2) = \frac{1}{T}\sum_{t=1}^{T}\|y_1^t - y_2^t\|_2.
\]
In this work, we consider the label/pure robustness with $D=\ADE$.
Note that other semantic metrics, such as metrics based on specific directions in \cite{zhang2022adversarial}, are also fully applicable to the above framework. In this work, we focus on robustness verification of trajectory prediction models:
\begin{quote}
Given a trajectory prediction model $g$, we determine whether $g$ is label-robust (or purely robust) at a given input $\hat {\mathbf X}$ w.r.t a given radius $r$.
\end{quote}

\section{Methodology}
\label{sec:model}

The most popular trajectory prediction models, including \cite{dendorfer2021mg,fang2020tpnet,sun2020reciprocal,chen2021personalized,liu2021social,salzmann2020trajectron++,yuan2021agentformer,cheng2021exploring,mangalam2020not,sun2021three,gu2022stochastic,xu2022remember}, are all very large with  stochastic output. As such, it is quite difficult to adopt traditional verification methods like SMT solving~\cite{katz2017reluplex,katz2019marabou} or abstract interpretation~\cite{singh2019abstract,yang2021improving} to verify their robustness properties. 

In \cite{li2022towards}, Li et al. proposed a black-box DNN verification algorithm \deeppac, where they relax the definition of robustness in a probabilistic way, allowing them to verify robustness at individual input regions using only a finite number of samples. This probably approximately correct (PAC) framework involves first learning a PAC model, an arbitrary function which (with probability close to $1$ at a given confidence level) approximates the DNN at the input region within a margin of discrepancy. Next, using this PAC model we can verify the robustness at the input region with guarantees on the confidence and error rate. 
Due to its black-box nature, \deeppac can be adapted to the robustness verification of trajectory prediction models. Moreover, the stochasticity of such models can be captured by PAC guarantees. We call our adapted method \trajpac, and in this section we detail how \trajpac is employed for robustness analysis of trajectory forecasting models.






\subsection{PAC Model Learning}
In Defs.~\ref{def:labelrobustness} and \ref{def:purerobustness}, the robustness of a trajectory prediction model requires the distance between the perturbed prediction and the ground truth/original prediction to be bounded by a safety constant. 
Thus, to analyse the label/pure robustness,
we learn a model approximating
the corresponding distance $D(Y,\cdot)$ 
with the PAC guarantee and further infer its maximal values.
Here we denote $\Delta(\mathbf X)$ as the distribution $D(g(\mathbf X),\cdot)$, where $\mathbf X=(X_0^\top,\ldots,X_N^\top)$ and $X_i \in B(\hat X_i,r)$ for each $i$.
Similar to \deeppac, we choose the function template to be an affine function, i.e,
$
\widetilde{\Delta}(\mathbf X)=\mathbf X \cdot \alpha + \beta,
$
where $\alpha$ and $\beta$ are constant real vectors which will be learned from sampling.
There are several reasons why we learn an affine function: 
First, the robustness properties we consider are all local robustness with a small neighboring region as the input region, and theoretically a continuous function can be approximated by an affine function with a very small error in a small region; after we learn the PAC model, we need to analyse how robust the PAC model is, and this analysis will be very easy and efficient if the PAC model is affine; also, an affine PAC model provides more accessible insight for model explanation.

To learn a function $\widetilde{\Delta}$ that fits $\Delta$ well, especially for a verification purpose, we desire that the difference of the two functions in the robustness region should be uniformly bounded by a margin $\lambda \ge 0$ as small as possible, so we have the following optimization problem:
\begin{equation}
    \label{dimensionlearn}
    \begin{array}{lll}
                & \min \lambda & \\
        \text{s.t.} & \sup_{d \in \Delta(\mathbf X)} |\widetilde\Delta(\mathbf X)-\Delta(\mathbf X)|\le \lambda ,\;\\
        & \forall \mathbf X \in B(X_0',r) \times \cdots \times B(X_N',r).
    \end{array}
\end{equation}
Generally it is difficult to solve \eqref{dimensionlearn}, since it has an uncountable number of constraints. Also, $\Delta(\mathbf X)$ is stochastic in nature, which makes solving this optimisation problem non-trivial.
Inspired by \deeppac and \cite{xue2020pac}, we can relax the problem \eqref{dimensionlearn} to finitely many constraints from the samples:
\begin{equation}
    \begin{array}{lll}
                & \min \lambda & \\
        \text{s.t.} & |\widetilde\Delta(\mathbf X)-\Delta(\mathbf X)|\le \lambda ,\;\\
        & \forall \mathbf X \in \mathcal X, \forall d \in \mathcal D(\mathbf X),
    \end{array}
\label{lplearn}
\end{equation}
where $\mathcal X \subseteq B(X_0',r) \times \cdots \times B(X_N',r)$ is a finite set of samples extracted independent and identically distributed from some distribution $\pi$, and $\mathcal D(\mathbf X) \subseteq D(g(\mathbf X),\cdot)$ is a finite set of samples from the distribution $D(g(\mathbf X),\cdot)$.
This relaxation is slightly different from that in \cite{li2022towards}, because we need to sample not only in the robustness region, but also in the distribution of the output distance $D(g(\mathbf X),\cdot)$.
The relaxed problem \eqref{lplearn} is a linear programming (LP), whose optimal can be obtained efficiently.
Since we only consider a finite subset of constraints, the optimal of \eqref{lplearn} does not necessarily satisfy all the constraints in \eqref{dimensionlearn}. In \cite{li2022towards}, a PAC guarantee can be constructed if we have enough samples, and we modify this result into our setting of trajectory prediction models, where stochastic output is considered.




\newtheorem{thm2}{Theorem}
\begin{thm2} \label{thm:main}
Let $\epsilon>0$ and $\eta>0$ be the pre-defined error rate and the significance level, respectively, and $K$ the number of samples. If
\begin{align} \label{eq:sample}
K \geq \frac{2}{\epsilon}\left(\ln \frac{1}{\eta}+2T_{\mathrm{p}}(N+1)+1\right),
\end{align}
then with confidence at least $1-\eta$, the optimal $\lambda^*$ of \eqref{lplearn} satisfies all the constraints in \eqref{dimensionlearn} but at most a fraction of probability $\epsilon$, i.e., $\mathbb{P}\left(|\widetilde\Delta(\mathbf X)-\Delta(\mathbf X)|\ge \lambda^* \right) \leq \epsilon$, where the probability measure $\mathbb P$ is the independent coupling of the sampling distribution $\pi$ and the random mechanism in the model $g(\cdot)$.
\end{thm2}


Thm.~\ref{thm:main} generalises the \deeppac method to trajectory prediction models, where we are faced with a regression model with random output, and different robustness properties. 
The essential difference is that the probability distribution $\mathbb P$, which is used for describing the PAC guarantee, is not  the sampling distribution, but its coupling with the random mechanism of the model. The proof of Thm.~\ref{thm:main} can be found in Appendix~\ref{appendix: Proof of Theorem 1}.

\begin{figure}[t]
\includegraphics[width=0.42
\textwidth]{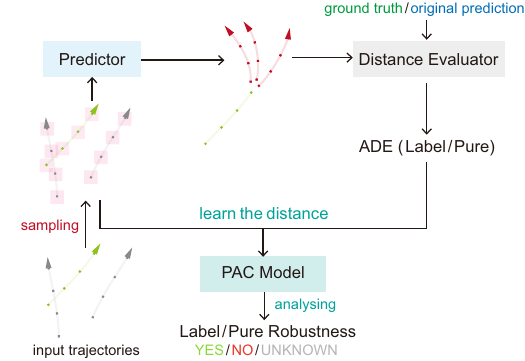}
\centering
\caption{Framework of robustness analysis}
\label{fig:framework}
\vspace{-1 em}
\end{figure}

Now our black-box framework of robustness analysis for trajectory prediction models is explicit, as is shown in \Cref{fig:framework}. Given the error rate $\epsilon$ and the significance level $\eta$, we extract $K$ samples in $B(X_0,r) \times \cdots \times B(X_N,r)$, where $K$ satisfies \eqref{eq:sample}. With the samples, we construct the linear programming problem \eqref{lplearn} and obtain (one of) its optimal, which gives the coefficients $\alpha$ and $\beta$ in the PAC model $\widetilde{\Delta}$ and the margin $\lambda^*$, and they will further help us analyse how robust the model is.

The optimisation of focused learning proposed in \cite{li2022towards} still fits in our settings, and we use it in our implementation. More details can be found in Appendix~\ref{appendix: focused learning}.


\subsection{Robustness Analysis}
We follow a similar robustness analaysis procedure to \deeppac.
 When the optimisation problem \eqref{lplearn} is solved,  we obtain the PAC model 
$\widetilde{\Delta}$ as well as the optimal margin $\lambda^*$. 
Intuitively, $\widetilde{\Delta}(\mathbf X) \pm \lambda^*$ approximates the upper/lower bound of $\Delta(\mathbf X)$ in the robustness region with the PAC guarantee. It is easy to see that, if the maximum of $\widetilde{\Delta}(\mathbf X) + \lambda^*$ is smaller than what the robustness property requires, i.e., the parameter $s$ in Def.~\ref{def:labelrobustness} or Def.~\ref{def:purerobustness}, then it holds under the same PAC guarantee. Since $\widetilde{\Delta}$ is an affine function, its maximum in a box region can be easily computed.

There are three circumstances that may occur in the robustness analysis:
\begin{itemize}
    \item The maximum of $\widetilde{\Delta}(\mathbf X) + \lambda^*$ is smaller than $s$. In this case, the robustness property holds with a PAC guarantee, and actually the models satisfies the so-call PAC-model robustness defined in \cite{li2022towards}. The analysis outputs YES. It is worth mentioning that, PAC-model robustness is far stronger than PAC robustness, especially that obtained from statistical methods like hypothesis testing or confidence interval calculation, because we have a model $\widetilde\Delta$ witnessing the robustness property PAC-true.
    \item The maximum of $\widetilde{\Delta}(\mathbf X) + \lambda^*$ is strictly larger than $s$, and we can find a true counterexample. In the model learning phase, if there exists a sample that violates the property, then it is a true counterexample. Also, when we calculate the maximum of $\widetilde{\Delta}(\mathbf X) + \lambda^*$, the maximum point $\arg \max \widetilde{\Delta}(\mathbf X)$ is likely to be a counterexample, and we run it in the original model to see whether it is a true counterexample. Once the PAC-model robustness does not hold, and we find a true counterexample in either way, the analysis outputs NO, i.e., the robustness property does not hold, with a true counterexample.
    \item It may occur that the maximum of $\widetilde{\Delta}(\mathbf X) + \lambda^*$ is strictly larger than $s$, but we cannot find a true counterexample. In this case, it is not sufficient to judge whether the model is robust or not according to the learned PAC model, so the analysis outputs UNKNOWN.
\end{itemize}
We remark that, in the first circumstance where the PAC-model robustness holds, we do not further check whether there is a true counterexample, because even if it exists, it does not violate the PAC-model robustness, in which the violation of the robustness property may occur with probability no more than the error rate $\epsilon$.


\subsection{Interpretation Analysis}

The PAC model we learn can also provide insight into two key features used by forecasting models when making predictions: the \textbf{critical paths} and the \textbf{critical steps} of agents. These two features are intuitive in the real world. For example, the movements of a person in front of you are more significant than the movements of someone behind you, and certain steps (e.g., changing direction) have greater impact than others.

As our PAC model is an affine function, 
there is a corresponding coefficient for every spatial coordinate in the input trajectory. 
The greater this coefficient magnitude is, the greater the impact of the corresponding coordinate's change on the prediction's label/pure ADE. We denote the $l^\infty$ normalized vector of coefficient magnitudes as the \emph{sensitivity} of our PAC model.


Therefore, spatial coordinates with high sensitivity values are identified as the critical steps, and trajectories with high average sensitivities are the critical paths. These critical steps and paths reflect which features in the historical trajectories can lead to vulnerabilities in the model. 
Not only does this give us a more interpretable understanding of how the model makes predictions, but it also allows us to analyse the key features that affect the model's robustness. Additionally, we can handcraft potential adversaries by only perturbing these key features, making our counterexamples highly intuitive.


\section{Experiments}
\label{sec: Experiment}

In this section, we evaluate our PAC-model robustness analysis method. We implement our algorithm \trajpac as a prototype. Its implementation is based on Python 3.7.8. 
Experiments are conducted on a Windows 11 PC with AMD R7, GTX 3070Ti, and 16G RAM. All the implementation and data
used in this section are publicly available\footnote {https://github.com/ZL-Helios/TrajPAC}.

\textbf{Datasets.}
We evaluate our method using the public
pedestrian trajectories forecasting benchmarks ETH/UCY ~\cite{pellegrini2010improving,lerner2007crowds} and the Stanford Drone Dataset (SDD)~\cite{robicquet2016learning}.  The ETH and UCY dataset group consists of five different scenes – ETH and HOTEL (from ETH), and UNIV, ZARA1 and ZARA2 (from UCY) and all the scenes report the position of pedestrians in world coordinates and hence the results are in metres. All the prediction models in our paper use the ``leave one-out" method~\cite{gupta2018social,huang2019stgat,kosaraju2019social,salzmann2020trajectron++} for training and evaluation. We follow the existing works that observing 8 frames (3.2 seconds) trajectories and predicting the next 12 frames (4.8 seconds). We randomly choose three predicted trajectories from each scenes for analysis, noted as (frame ID , person ID).
Experiments regarding SDD can be found in Appendix~\ref{appendix: Experiments on SDD}.

\textbf{Prediction Models.}
In our paper, we analyse four state-of-art multi-model prediction models: Trajectron++~\cite{salzmann2020trajectron++}, AgentFormer~\cite{yuan2021agentformer}, MemoNet~\cite{xu2022remember} and MID~\cite{gu2022stochastic}.

\textbf{Sampling.}
The sampling distribution $\pi$ is the uniform distribution on the robustness region. 
When we calculate the ADE of the samples, we use a modified version of ADE, the minimum average displacement error of $K$ trajectory samples, 
which is a standard metric for trajectory prediction \cite{gupta2018social,sadeghian2019sophie,salzmann2020trajectron++,phan2020covernet,chai2019multipath}.
We claim that this will not break the PAC guarantee in Thm.~\ref{thm:main}. More details can be found in Appendix~\ref{appendix: Robustness Properties for ADE}. In our experiment, we choose $K = 20$.

\textbf{Implementation details.}
In the later part, we choose 1 meter/0.5 meters to be the safety constant for label/pure robustness analysis with the perturbation
radius $r = 0.03$ meters, respectively. Experiments with varying values of $r$ can be found in Appendix~\ref{appendix: Experiments on r}. As for PAC model learning, we choose $\eta = 0.01$ and $\epsilon = 0.01$.

In what follows, we are going to answer the research questions below:

 \noindent {\bf RQ1:} Does \trajpac perform well in verifying robustness?
 
 
 \noindent {\bf RQ2:} Can \trajpac precisely capture the robustness performance of the prediction models?
 

 \noindent {\bf RQ3:} Can \trajpac provide intuitive analysis of the robustness performance of different prediction models?

\subsection{Robustness Analysis of Different Models}

First, we evaluate the performance of \trajpac on giving robustness verification.
This includes whether the model can achieve robustness prediction for a given safety constant and perturbation radius, whether it can be applied to a wide range of models with high validation efficiency, and whether those cases verified as robust demonstrate good anti-attack performance.

\begin{table}[t]
\renewcommand\arraystretch{0.95}
\scalebox{0.53}{
\begin{tabular}{ccrcccccccccc}
\specialrule{.08em}{.0em}{.0em} 
\multirow{2}{*}{Scene} && \multicolumn{1}{c}{\multirow{2}{*}{ID}} && \multicolumn{4}{c}{Label Robustness}  && \multicolumn{4}{c}{Pure Robustness} \\ 
 && && Traj++ & Memo & AgentF & MID  && Traj++ & Memo & AgentF & MID \\ \cline{1-1} \cline{3-3} \cline{5-8} \cline{10-13}
 && (4400, 79) && {\color[HTML]{F8A102} $\ocircle$} & {\color[HTML]{009901} \cmark} & {\color[HTML]{CB0000} \xmark} & {\color[HTML]{CB0000} \xmark}  && {\color[HTML]{F8A102} $\ocircle$} & {\color[HTML]{009901} \cmark} & {\color[HTML]{CB0000} \xmark} & {\color[HTML]{009901} \cmark} \\
 && (6490, 127) && {\color[HTML]{009901} \cmark} & {\color[HTML]{009901} \cmark} & {\color[HTML]{CB0000} \xmark} & {\color[HTML]{CB0000} \xmark}  && {\color[HTML]{009901} \cmark} & {\color[HTML]{009901} \cmark} & {\color[HTML]{CB0000} \xmark} & {\color[HTML]{009901} \cmark} \\
\multirow{-3}{*}{ETH} && (10340, 257) && {\color[HTML]{F8A102} $\ocircle$} & {\color[HTML]{F8A102} $\ocircle$} & {\color[HTML]{CB0000} \phantom{$^\dag$}\xmark$^\dag$} & {\color[HTML]{CB0000} \xmark} && {\color[HTML]{F8A102} $\ocircle$} & {\color[HTML]{009901} \cmark} & {\color[HTML]{CB0000} \xmark} & {\color[HTML]{CB0000} \xmark} \\ \cline{1-1} \cline{3-3} \cline{5-8} \cline{10-13}
 && (7550, 157) && {\color[HTML]{009901} \cmark} & {\color[HTML]{009901} \cmark} & {\color[HTML]{CB0000} \xmark} & {\color[HTML]{F8A102} $\ocircle$} && {\color[HTML]{009901} \cmark} & {\color[HTML]{009901} \cmark} & {\color[HTML]{009901} \cmark} & {\color[HTML]{009901} \cmark} \\
 && (10530, 236) && {\color[HTML]{009901} \cmark} & {\color[HTML]{009901} \cmark} & {\color[HTML]{F8A102} $\ocircle$} & {\color[HTML]{009901} \cmark} && {\color[HTML]{009901} \cmark} & {\color[HTML]{009901} \cmark} & {\color[HTML]{009901} \cmark} & {\color[HTML]{009901} \cmark} \\
\multirow{-3}{*}{Hotel} && (15030, 345) && {\color[HTML]{009901} \cmark} & {\color[HTML]{009901} \cmark} & {\color[HTML]{009901} \cmark} & {\color[HTML]{009901} \cmark} && {\color[HTML]{009901} \cmark} & {\color[HTML]{009901} \cmark} & {\color[HTML]{009901} \cmark} & {\color[HTML]{009901} \cmark} \\ \cline{1-1} \cline{3-3} \cline{5-8} \cline{10-13}
 && (4430, 69) && {\color[HTML]{F8A102} $\ocircle$} & {\color[HTML]{009901} \cmark} & {\color[HTML]{CB0000} \xmark} & {\color[HTML]{CB0000} \xmark} && {\color[HTML]{009901} \cmark} & {\color[HTML]{009901} \cmark} & {\color[HTML]{009901} \cmark} & {\color[HTML]{009901} \cmark} \\
 && (6050, 102) && {\color[HTML]{009901} \cmark} & {\color[HTML]{009901} \cmark} & {\color[HTML]{CB0000} \xmark} & {\color[HTML]{CB0000} \xmark} && {\color[HTML]{009901} \cmark} & {\color[HTML]{009901} \cmark} & {\color[HTML]{F8A102} $\ocircle$} & {\color[HTML]{009901} \cmark} \\
\multirow{-3}{*}{Zara1} && (8680, 142) && {\color[HTML]{CB0000} \phantom{$^\dag$}\xmark$^\dag$} & {\color[HTML]{F8A102} $\ocircle$} & {\color[HTML]{CB0000} \phantom{$^\dag$}\xmark$^\dag$} & {\color[HTML]{009901} \cmark} && {\color[HTML]{F8A102} $\ocircle$} & {\color[HTML]{009901} \cmark} & {\color[HTML]{009901} \cmark} & {\color[HTML]{009901} \cmark} \\ \cline{1-1} \cline{3-3} \cline{5-8} \cline{10-13}
 && (3400, 65) && {\color[HTML]{009901} \cmark} & {\color[HTML]{009901} \cmark} & {\color[HTML]{CB0000} \xmark} & {\color[HTML]{009901} \cmark} && {\color[HTML]{009901} \cmark} & {\color[HTML]{009901} \cmark} & {\color[HTML]{009901} \cmark} & {\color[HTML]{009901} \cmark} \\
 && (7430, 141) && {\color[HTML]{009901} \cmark} & {\color[HTML]{009901} \cmark} & {\color[HTML]{CB0000} \xmark} & {\color[HTML]{CB0000} \xmark} && {\color[HTML]{009901} \cmark} & {\color[HTML]{009901} \cmark} & {\color[HTML]{F8A102} $\ocircle$} & {\color[HTML]{F8A102} $\ocircle$} \\
\multirow{-3}{*}{Zara2} && (10030, 195) && {\color[HTML]{CB0000} \xmark} & {\color[HTML]{009901} \cmark} & {\color[HTML]{CB0000} \xmark} & {\color[HTML]{CB0000} \xmark} && {\color[HTML]{CB0000} \xmark} & {\color[HTML]{F8A102} $\ocircle$} & {\color[HTML]{F8A102} $\ocircle$} & {\color[HTML]{009901} \cmark} \\ \cline{1-1} \cline{3-3} \cline{5-8} \cline{10-13}
 && (1840, 105) && {\color[HTML]{CB0000} \xmark} & {\color[HTML]{CB0000} \xmark} & {\color[HTML]{CB0000} \phantom{$^\dag$}\xmark$^\dag$} & {\color[HTML]{CB0000} \xmark} && {\color[HTML]{F8A102} $\ocircle$} & {\color[HTML]{009901} \cmark} & {\color[HTML]{F8A102} $\ocircle$} & {\color[HTML]{009901} \cmark} \\
 && (4820, 202) && {\color[HTML]{CB0000} \phantom{$^\dag$}\xmark$^\dag$} & {\color[HTML]{CB0000} \xmark} & {\color[HTML]{CB0000} \phantom{$^\dag$}\xmark$^\dag$} & {\color[HTML]{F8A102} $\ocircle$} && {\color[HTML]{F8A102} $\ocircle$} & {\color[HTML]{009901} \cmark} & {\color[HTML]{F8A102} $\ocircle$} & {\color[HTML]{009901} \cmark} \\
\multirow{-3}{*}{Univ} && (5250, 297) && {\color[HTML]{009901} \cmark} & {\color[HTML]{009901} \cmark} & {\color[HTML]{CB0000} \xmark} & {\color[HTML]{009901} \cmark} && {\color[HTML]{F8A102} $\ocircle$} & {\color[HTML]{009901} \cmark} & {\color[HTML]{F8A102} $\ocircle$} & {\color[HTML]{009901} \cmark} \\ \specialrule{.08em}{.0em}{.0em} 
\end{tabular}
}
\centering
\caption{Label/pure robustness verification.\; We mark {\color[HTML]{009901}\cmark} if it is PAC-model robust, i.e., the robustness analysis returns YES, {\color[HTML]{CB0000}\xmark} if the PAC-model with the optimal margin is not robust and we find a true counterexample, i.e., the robustness analysis returns NO, and {\color[HTML]{F8A102}$\ocircle$} otherwise, i.e., the robustness analysis returns UNKNOWN.
We use $\dag$ to indicate that PGD attacks successfully, i.e., the adversary found by PGD exceeds the robustness threshold.}
\label{table: define}
\end{table}

As shown in \Cref{table: define}, as a black-box method, \trajpac can analyse label and pure robustness of different trajectory prediction models, showing good scalability. The sampling time varies among different prediction models though, yet time for PAC-model learning and robustness analysis is quite short, as shown in \Cref{tabel: time}. This demonstrates that \trajpac is very efficient in analysing large trajectory prediction models.

\begin{table}[t]
\resizebox{\columnwidth}{!}{
\begin{tabular}{ccrrrrrcc}
\specialrule{.12em}{.0em}{.05em}
\multicolumn{1}{c}{\multirow{2}{*}{Method}} && \multicolumn{5}{c}{Average Sampling Rate (iteration/s)} && \multirow{2}{*}{\begin{tabular}[c]{@{}c@{}}Average PAC-Model\\ Learning Time (s)\end{tabular}} \\ \cline{3-7}
 && \multicolumn{1}{l}{ETH} & Hotel & Zara1 & Zara2 & Univ &  \\\cline{1-1} \cline{3-7} \cline{9-9} 
Traj++ && 51.50 & 52.15 & 52.27 & 51.87 & 52.26 && 1.02 \\
MemoNet && 0.99 & 1.92 & 1.91 & 1.32 & 0.94 && 1.05 \\
AgentFormer && 15.57 & 16.22 & 16.30 & 15.38 & 12.38 && 1.10 \\
MID && 0.14 & 0.13 & 0.13 & 0.14 & 0.13 && 0.15 \\
\specialrule{.12em}{.0em}{.0em}
\end{tabular}
}
\centering
\caption{The average sampling rate, in iterations per second, of each model at each scene. The diffusion-based model (MID) has the longest sampling rate, in which 10000 samples require a time of $\sim$20 hours. Because of this, we opt to use fewer samples for its scenario optimization process, resulting in the faster PAC learning time.}
\label{tabel: time}
\end{table}

\begin{table}[t]
\renewcommand\arraystretch{0.95}
\scalebox{0.82}{
\begin{tabularx}{0.578\textwidth}{ccXXXX}
\specialrule{.1em}{0pt}{0pt}
\multicolumn{1}{c}{\multirow{2}{*}{Scene}} && \multicolumn{4}{c}{ADE$_{20}$ (in metre), best-of-20 samples} \\ \cline{3-6} 
 && Traj++ & MemoNet & AgentFormer & \quad MID \\ \cline{1-1} \cline{3-6}
ETH && \,0.46 & 0.41 & 0.41 & \quad 0.51 \\
Hotel && \,0.15 & 0.14 & 0.30 & \quad 0.15 \\
Zara1 && \,0.49 & 0.57 & 0.34 & \quad 0.25 \\
Zara2 && \,0.36 & 0.33 & 0.27 & \quad 0.27 \\
Univ && \,0.69 & 0.54 & 0.62 & \quad 0.31 \\
\cline{1-1} \cline{3-6}
Average ADE && \,0.43 & 0.40 & 0.39 & \quad 0.30 \\ \specialrule{.1em}{0em}{0em}
\end{tabularx}
}
\caption{
Average predicted $\mathrm{ADE}_{20}$ scores for the three verification samples per scene.}
\label{table: Average ADE}
\vspace{-1 em}
\end{table}

\trajpac only provides a PAC guarantee, so we are concerned with the soundness of its robustness analysis. We conduct PGD attacks on Trajectron++ and AgentFormer like \cite{cao2022advdo}. For the cases \trajpac outputs YES, PGD does not find any true counterexamples, which implies that \trajpac is sound empirically. \trajpac is conservative in analysing robustness: Even on the UNKNOWN cases, there is no successful PGD attack. Also, \trajpac shows good performance in finding counterexamples, as \trajpac can find counterexamples to which PGD does not get access.

Due to the model performance on motion forecasting affecting its label robustness, we provide the predicted ADE in \Cref{table: Average ADE}.
The four methods show relatively similar average performance without perturbation, so in terms of maintaining accurate predictions in the face of perturbation, Memonet does exhibit stronger label robustness, as emphasised in \cite{xu2022remember}. Similar to the findings in \cite{cao2022advdo}, Trajectron++ exhibits stronger label robustness compared to Agentformer. We can see that the analysis results given by \trajpac are consistent with other methods.

\vspace{1mm}
\noindent\doublebox{
\begin{minipage}{0.94\linewidth}
{\bf Answer RQ1:} \trajpac shows good scalability, efficiency and soundness in robustness analysis of different trajectory prediction models. Its results of robustness analysis are consistent with other methods.
\end{minipage}
}

\subsection{Precision of the PAC models}

\begin{figure*}[t]
\includegraphics[width=\textwidth]{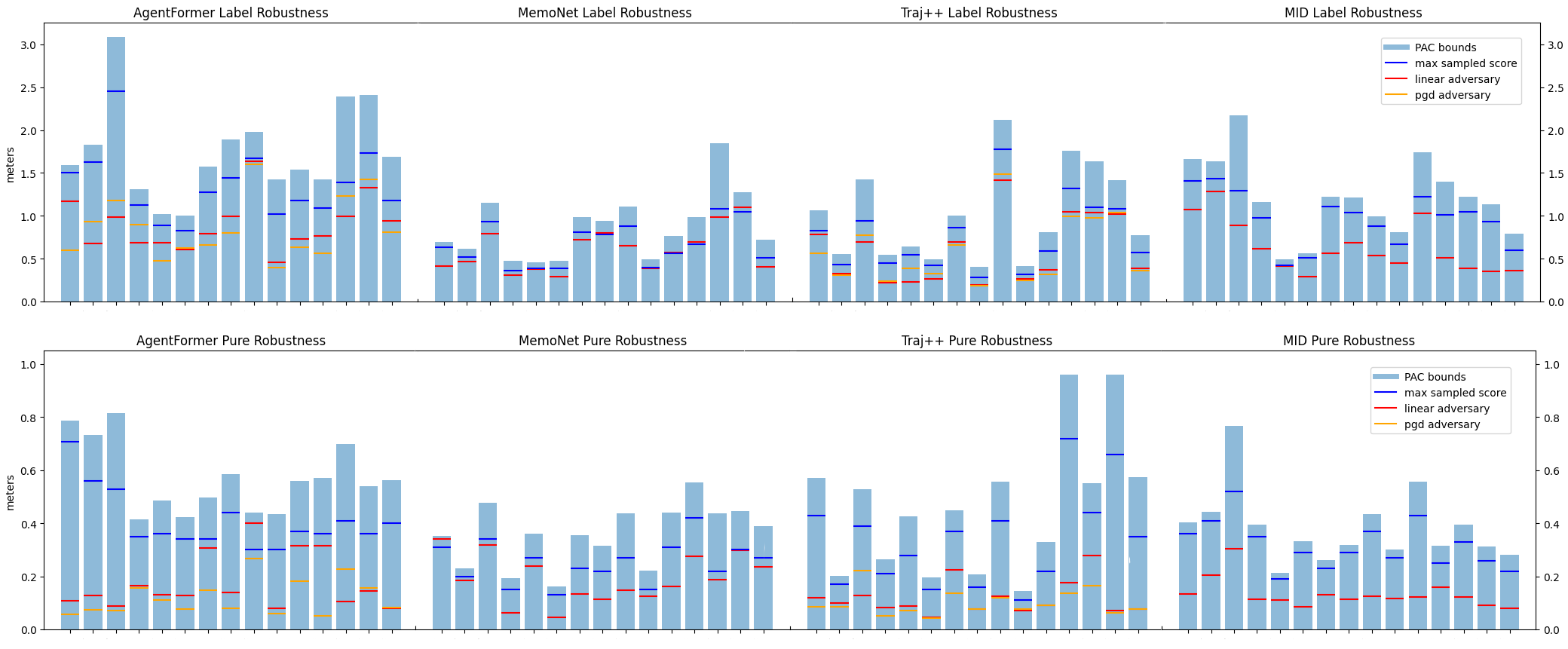}
\caption{Visualizations of the PAC ADE upper bounds (blue bars), maximum sampled ADE encountered in the PAC model learning process (blue stripes), and ADE of linear and pgd adversaries (red and orange stripes respectively) from our PAC model and PGD attacks.}
\label{fig: ADE bounds}
\end{figure*}

\begin{figure*}[t]
\includegraphics[width=0.95\textwidth]{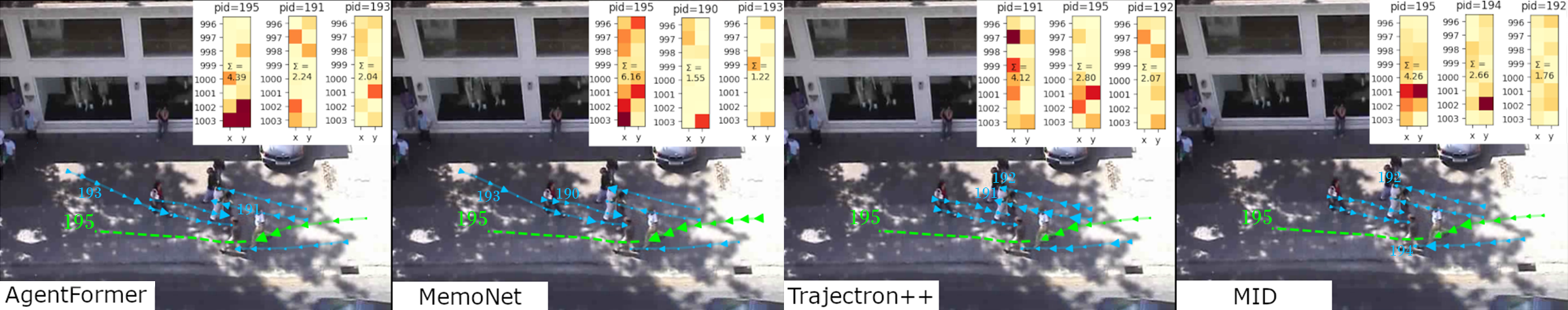}
\centering
\caption{
Sensitivity plots for each prediction model at sample (10030, 195) from scene Zara2. The green path is the agent trajectory and the blue paths are neighboring trajectories. The size of the directional arrows are proportional to the sensitivity of our PAC linear model at that position. The top right of each plot contains a heatmap of the top three critical paths. Darker colors in the heatmap represent higher sensitivity values. The value inside each heatmap is the sum of all sensitivities in the path.
}
\label{fig: interpretation}
\vspace{-1 em}
\end{figure*}

Generally it is difficult to straight evaluate how precise the PAC models learned by \trajpac are. Here we calculate four ADE estimations, namely
the ADE upper bound given by the PAC model, the ADE of the adversary generated by our PAC model, the maximum ADE among the samples required for training \trajpac, and the ADE of the adversary from PGD attack; the first two are estimations from the PAC model, while the latter two are ADE performance of the model in the robustness region. In \Cref{fig: ADE bounds}, we present a detailed illustration of the four estimations.

\begin{table}[t]
\resizebox{\columnwidth}{!}{
\begin{tabular}{cccccc}
\specialrule{.1em}{.03em}{.03em}
\multicolumn{1}{c}{\multirow{2}{*}{Scene}} && \multicolumn{4}{c}{\begin{tabular}[c]{@{}c@{}}PAC ADE upper bound - max sampled ADE\\ (label robustness / pure robustness)\end{tabular}} \\ \cline{3-6} 
&& Traj++ & MemoNet & AgentFormer & MID \\ \cline{1-1} \cline{3-6}
ETH && 0.28 / 0.10 & 0.13 / 0.07 & 0.31 / 0.18 & 0.45 / 0.11 \\
Hotel && 0.08 / 0.08 & 0.09 / 0.06 & 0.16 / 0.09 & 0.10 / 0.04 \\
Zara1 && 0.21 / 0.09 & 0.19 / 0.13 & 0.35 / 0.15 & 0.13 / 0.04 \\
Zara2 && 0.25 / 0.13 & 0.20 / 0.11 & 0.37 / 0.18 & 0.35 / 0.07 \\
Univ && 0.36 / 0.21 & 0.40 / 0.16 & 0.73 / 0.21 & 0.19 / 0.06 \\ \cline{1-1} \cline{3-6}
Average && 0.24 / 0.12 & 0.20 / 0.11 & 0.38 / 0.16 & 0.24 / 0.06 \\ 
\specialrule{.1em}{.0em}{.0em}
\end{tabular}}
\caption{
Differences between our computed PAC ADE upper
bound and the maximum sampled ADE during the model learning process.
}
\label{table:upperbound-max}
\end{table}

\begin{table}[t]
\renewcommand\arraystretch{0.95}
\centering
\scalebox{0.8}{
\begin{tabularx}{0.59\textwidth}{ccXX}
\specialrule{.1em}{.0em}{.05em}
\multicolumn{1}{c}{\multirow{2}{*}{Scene}} && \multicolumn{2}{c}{\begin{tabular}[c]{@{}c@{}} $\lvert\text{ADE}_\text{pgd} - \text{ADE}_\text{linear}\rvert$ \\ (label robustness / pure robustness)\end{tabular}} \\ \cline{3-4} 
 && \qquad \qquad Traj++ & \quad AgentFormer \\ \cline{1-1} \cline{3-4}
ETH && \qquad \quad 0.10 / 0.05 & \quad \; 0.34 / 0.04 \\
Hotel && \qquad \quad 0.08 / 0.02 & \quad \; 0.15 / 0.03 \\
Zara1 && \qquad \quad 0.04 / 0.03 & \quad \; 0.12 / 0.12 \\
Zara2 && \qquad \quad 0.05 / 0.02 & \quad \; 0.12 / 0.14 \\
Univ && \qquad \quad 0.04 / 0.04 & \quad \; 0.16 / 0.05 \\ \cline{1-1} \cline{3-4}
Average && \qquad \quad 0.06 / 0.03 & \quad \; 0.18 / 0.08 \\ 
\specialrule{.1em}{.0em}{.0em}
\end{tabularx}}
\caption{\centering ADE of adversaries from PAC models and PGD.}
\label{tabel:pgd-linear} 
\vspace{-1em}
\end{table}

First, the ADE upper bound given by \trajpac is very close to (and still above) the maximum sampled ADE during the model learning process, indicating that our PAC model has captured the behavior of the original model well and produced highly  accurate ADE upper bounds, as shown in \Cref{table:upperbound-max}.  
Furthermore, the robustness analysis results obtained through our method still exhibit significant soundness in \Cref{fig: ADE bounds}, as the ADE of the linear adversary generated from PAC model, as well as PGD adversary, are all smaller than the ADE upper bound.
 Also, we notice that the adversaries generated by our PAC model are as effective as PGD, since the ADE of the adversary generated by our PAC model is very close to that of PGD adversary, as is depicted in \Cref{tabel:pgd-linear}. From \Cref{fig: ADE bounds}, the adversaries generated by our method exhibit better overall attack effectiveness compared to PGD in the analysis of pure robustness.

\vspace{1mm}
\noindent\doublebox{
\begin{minipage}{0.94\linewidth}
{\bf Answer RQ2:} \trajpac can provide tight ADE upper bound of different prediction methods.  Adversaries generated from \trajpac exhibit comparable (and even better) performance to adversaries found by PGD. 
\end{minipage}
}

\subsection{Interpretation Analysis}

We perform an interpretation analysis on the sample (10030, 195) in Zara2. Among the four methods, MemoNet is the only label-robust method, with a label ADE upper bound of $0.98$. Trajectron++ is the least robust, with an upper bound of 1.76. In \Cref{fig: interpretation} we visualise the critical steps of different prediction methods, and shows the top three critical paths in each method. Based on \Cref{fig: interpretation}, we emphasise the following observations: Steps closer to the present are more likely to be critical steps, and the trajectory of the agent itself is often the critical path. 

Our analysis also exposes potential vulnerabilities at each sample. For instance, the critical paths captured by MemoNet (190 and 193) are walking directly towards the agent, whereas the critical paths captured by Trajectron++ (191 and 192) are walking away. Knowing this, black-box attackers are able to handcraft adversaries by adding perturbations to only these key positions. In particular,  the critical paths of Trajectron++ makes it more susceptible to attacks, since defenses are more likely to focus on the paths walking directly towards the agent, rather than those walking away. 

\vspace{1mm}

\noindent\doublebox{
\begin{minipage}{0.94\linewidth}

{\bf Answer RQ3:} \trajpac can identify key features that contribute to the overall performance and robustness through sensitivity analysis of the PAC model.
\end{minipage}
}

\section{Conclusion}

We present \trajpac for robustness verification of trajectory prediction models. It is highly scalable, efficient, empirically sound, and capable of generating adversaries and interpretation. As for future works, we will consider more realistic safety properties in trajectory prediction, and how to use the verification results of trajectory prediction to analyse the safety of autonomous driving scenarios.


\section*{Acknowledgements}
Supported by CAS Project for Young Scientists in Basic Research, Grant No.YSBR-040, and ISCAS New Cultivation Project  ISCAS-PYFX-202201.

{\small
\bibliographystyle{ieee_fullname}
\bibliography{egbib}
}

\newpage
\appendix
\onecolumn


\section{Proof of Theorem 1}
\label{appendix: Proof of Theorem 1}
To prove Thm.~\ref{thm:main}, we first introduce the theory of scenario optimisation. Let us take a look at the optimization problem below: 
\begin{equation}
    \label{robustopt}
    \begin{split}
        & \min\limits_{\boldsymbol{\gamma}\in \Gamma \subseteq \mathbb{R}^m}\boldsymbol{b}^\top\boldsymbol{\gamma},\\
        s.t.\;& f_{\boldsymbol{\omega}}(\boldsymbol{\gamma})\leq 0,\;\forall \boldsymbol{\omega}\in \Omega ,
    \end{split}
\end{equation}
where $f_{\boldsymbol{\omega}}$ is a convex and continuous function of the $m$-dimensional optimization variable $\boldsymbol{\gamma}$ for every $\boldsymbol{\omega}\in \Omega$, and both $\Omega$ and $\Gamma$ are convex and closed.
It is difficult to solve \eqref{robustopt}, since there are infinitely many constraints.
In \cite{DBLP:journals/tac/CalafioreC06}, Calafiore et al. proposed the following scenario optimisation to solve \eqref{robustopt} with a PAC guarantee.

\begin{thm}
    \label{scenariodef}
   Let $\mathbb P$ be a probability measure on $\Omega$. The scenario approach to handle the optimization problem~\eqref{robustopt} is to solve the following problem. We extract $K$ independent and identically distributed (i.i.d.) samples $(\boldsymbol{\omega}_i)_{i=1}^K$ from $\Omega$ according to the probability measure $\mathbb P$:
    \begin{equation}
    \label{scenarioopt}
    \begin{split}
        & \min\limits_{\boldsymbol{\gamma}\in \Gamma \subseteq \mathbb{R}^m}\boldsymbol{b}^\top\boldsymbol{\gamma},\\
        \mathrm{s.t.}\;&\bigwedge_{i=1}^{K} f_{\boldsymbol{\omega}_i}(\boldsymbol{\gamma})\leq 0.
    \end{split}
\end{equation}
\end{thm}
The scenario optimisation only considers a finite subset of constraints. In \cite{DBLP:journals/tac/CalafioreC06,DBLP:journals/arc/CampiGP09}, a PAC guarantee
between the scenario solution in \eqref{scenarioopt} and its original optimization in \eqref{robustopt} can be constructed with sufficient samples.

\begin{thm2}[\cite{DBLP:journals/arc/CampiGP09}]\label{scenariotheorem}
    If (\ref{scenarioopt}) is feasible and has an optimal solution $\boldsymbol{\gamma}^*_K$, and
    \begin{equation}
        \label{eq:scenariotheorem}
        \epsilon\geq \frac{2}{K}(\ln\frac{1}{\eta}+m),
    \end{equation}
    where $K$ is the number of samples, and $\epsilon$ and $\eta$ are the pre-defined error rate and the significance level, respectively, then with confidence at least $1-\eta$, the optimal $\boldsymbol{\gamma}^*_K$ satisfies all the constraints in $\Omega$ but only at most a fraction of probability measure $\epsilon$, i.e., $\mathbb P (f_{\boldsymbol{\omega}}(\boldsymbol{\gamma}_K^*)> 0)\leq \epsilon$.
\end{thm2}

In \deeppac~\cite{li2022towards}, scenario optimisation is used for robustness verification of classification DNNs. 
To adapt Thm.~\ref{scenariotheorem} to our settings, where stochastic output is considered, we must describe both the sampling distribution $\pi$ and the stochasticity in the trajectory prediction model in the probability distribution $\mathbb P$. 

For label robustness, stochasticity in $\Delta(\mathbf X)$ only comes from the stochasticity of the output  $g(\mathbf X)$. We regard the model $g$ as a random variable $g(\mathbf X,\omega): (\varOmega,\mathcal F,\Pr_{\mathbf X}) \to (\mathbb R^{2\times T_\f},\mathcal B(\mathbb R^{2\times T_\f}))$, where $(\varOmega,\mathcal F,\Pr_{\mathbf X})$ is the probability space of the stochasticity in $g(\mathbf X)$, and $\mathcal B(\cdot)$ is the Borel $\sigma$-algebra, i.e., the $\sigma$-algebra generated by the open sets. 
Now we consider the product measurable space $(B(\hat{\mathbf X},r) \times \varOmega,\mathcal B(B(\hat{\mathbf X},r)) \times \mathcal F)$ and we define the probability measure $\mathbb P$ on it according to the sampling distribution $\pi$ and the probability measure $\Pr_{\mathbf X}$ in the standard way: For $B \in \mathcal B(B(\hat{\mathbf X},r))$ and $F \in \mathcal F$, the measure of the measurable rectangle $B \times F$ is 
$$
\mathbb P(B \times F) = \int_B \Pr\nolimits_{\mathbf X}(F)  \pi(\dif \mathbf X);
$$
it is easy to see that $\mathbb P$ is a probability measure on the semi-ring of the measurable rectangles in $\mathcal B(B(\hat{\mathbf X},r)) \times \mathcal F$, and thus it can be uniquely extended to a probability measure, still denoted by $\mathbb P$, on $(B(\hat{\mathbf X},r) \times \varOmega,\mathcal B(B(\hat{\mathbf X},r)) \times \mathcal F)$. When sampling in $B(\hat{\mathbf X},r)$ according to $\pi$, we are actually sampling in the probability space 
$(B(\hat{\mathbf X},r) \times \varOmega,\mathcal B(B(\hat{\mathbf X},r)) \times \mathcal F,\mathbb P)$, so according to Thm.~\ref{scenariotheorem}, where the dimensionality $m=2T_{\mathrm{p}}(N+1)+1$, it suffices to prove Thm.~\ref{thm:main}. 

For pure robustness, stochasticity in $\Delta(\mathbf X)$ comes from the stochasticity of both $g(\mathbf X)$ and $g(\hat{\mathbf X})$, so we sample in the measurable space $(B(\hat{\mathbf X},r) \times \varOmega \times \varOmega,\mathcal B(B(\hat{\mathbf X},r)) \times \mathcal F \times \mathcal F)$, and the probability of a measurable rectangle $B \times F_1 \times F_2$ is
$$
\mathbb P(B \times F_1 \times F_2) = \int_B \Pr\nolimits_{\mathbf X}(F_1)  \pi(\dif \mathbf X) \cdot \Pr\nolimits_{\hat{\mathbf X}}(F_2).
$$
By measure extension, $\mathbb P$ is a probability measure on $(B(\hat{\mathbf X},r) \times \varOmega \times \varOmega,\mathcal B(B(\hat{\mathbf X},r)) \times \mathcal F \times \mathcal F)$. With the same dimensionality $m=2T_{\mathrm{p}}(N+1)+1$, Thm.~\ref{thm:main} is proved.

The deduced PAC-model robustness is obviously for label robustness: When $\max_{\mathbf X \in B(\hat{\mathbf X},r)}\widetilde\Delta(\mathbf X)+\lambda^* \le s$, with confidence $1-\eta$ we have 
    \[
    \begin{aligned}
 \mathbb{P}({\Delta}(\mathbf{X})  \leq s)
    \geq \mathbb{P}({\Delta}(\mathbf{X}) 
    \leq {\widetilde\Delta}(\mathbf{X})+\lambda^*)
     \geq\mathbb{P}(|{\widetilde\Delta}(\mathbf{X})-{\Delta}(\mathbf{X})|\leq \lambda^*)
    \geq 1-\epsilon,
        \end{aligned}
\]
which implies that $g$ is PAC-model label-robust in $B({\hat{\mathbf{X}}},r)$. As for pure robustness, with the same deduction, when $\max_{\mathbf X \in B(\hat{\mathbf X},r)}\widetilde\Delta(\mathbf X)+\lambda^* \le s$, with confidence $1-\eta$ we have $ \mathbb{P}({\Delta}(\mathbf{X})  \leq s) \geq 1-\epsilon$, indicating that with a PAC guarantee, $D(Y,\hat Y) \le s$ holds for any $Y \in g(\mathbf X)$ and any $\hat Y \in g(\hat{\mathbf X})$, which is a stronger property than the pure robustness defined in Def.~\ref{def:purerobustness}.

It is worth mentioning that, even if we use focused learning detailed in Appendix \ref{appendix: focused learning}, the PAC guarantee given by Thm.~\ref{thm:main} will not be violated, since the PAC guarantee is constructed only in the second learning phase, while we only obtain an affine function template with fewer coefficients to be determined in the first learning phase.

\section{Focused Learning}
\label{appendix: focused learning}

We employ a focused learning procedure for PAC model learning first described in \cite{li2022towards}. The basic idea involves splitting the model learning stage into two, more manageable, subphases. The first subphase involves extracting $\mathcal{K}$ key features from the model based on the $\mathcal{K}$ largest coefficient magnitudes. In the second subphase we optimize our PAC model with respect to only those previously found key features. The main idea of this procedure is outlined below:
\begin{enumerate}
    \item \emph{First learning phase}: We learn the scores, i.e., ADEs, $\Delta_{t=1..T_1}$ for $T_1$ i.i.d. samples from the input region $B(\hat{x}, r)$. This LP problem has $d$ variables with $T_1$ constraints. For large datasets this LP problem is still too large, and so we can instead use linear regression to boost the learning time. After solving the linear problem, we find the $\mathcal{K}$ largest coefficient magnitudes, and denote the set of corresponding features by $Key(\mathcal{K}) \subseteq \{1, x_1, ..., x_d\}$.
    \item \emph{Second learning phase}: We learn the scores $\Delta_{t=1..T_2}$ for $T_2$ i.i.d. samples from $B(\hat{x}, r)$. Rather than solving an LP problem for all $d$ variables, we fix the non-key coefficients and generate constraints for only our $\mathcal{K}$ key features. The solution to this LP problem determines the coefficients of these key features $Key(\mathcal{K})$.
\end{enumerate}

\noindent
With focused learning, rather than optimizing a large LP problem with $d$ variables and $T_1 + T_2$ constraints, we solve only one LP problem with $\mathcal{K} \leq d$ variables and $T_2$ constraints. Moreover, given a predetermined significance $\eta$ and error rate $\epsilon$, we can determine an appropriate number of key features $\mathcal{K}$ and sample size $T_2$ satisfying $\mathcal{K} \leq \frac{\epsilon T_2}{2} - \ln \frac{1}{\eta} - 1$ \cite[Theorem~2.5]{li2022towards}. 

\begin{table}[ht]
\centering
\begin{tabular}{clllll}
\hline
 & \multicolumn{1}{c}{$r$} & \multicolumn{1}{c}{$\epsilon$} & \multicolumn{1}{c}{$\eta$} & \multicolumn{1}{c}{$T_1$} & \multicolumn{1}{c}{$T_2$} \\ \hline
Traj++ & \multirow{4}{*}{0.03} & \multirow{4}{*}{0.01} & \multirow{4}{*}{0.01} & 30000 & 12000 \\
MemoNet &  &  &  & 20000 & 12000 \\
AgentFormer &  &  &  & 30000 & 12000 \\
MID &  &  &  & 4000 & 3000 \\ \hline
\end{tabular}
\caption{Detailed hyperparameter configurations for scenario optimization of each trajectory forecasting model.}
\end{table}

\section{Robustness Properties for $\mathrm{ADE}_K$}
\label{appendix: Robustness Properties for ADE}

In our experiment, we use a modified version of ADE, the minimum average displacement error of $K$ trajectory samples, 
which is a standard metric for trajectory prediction \cite{gupta2018social,sadeghian2019sophie,salzmann2020trajectron++,phan2020covernet,chai2019multipath}.
Formally, it is defined as
 \[
 \ADE_K(\mathbb{Y}, Y) = \frac{1}{T}\min\limits _{1 \le k \le K}\sum_{t=1}^{T}\|y^{t,(k)} - y^t\|_2,
 \]
 where $\mathbb{Y}=\{(y^{1,(k)},\ldots,y^{T,(k)} ) \mid k=1,\ldots,K\}$ is a set of $K$ trajactory samples, and $y^{t,(k)}$ is the position at time $t$ in the $k$-th sample. 
In the experiment, we choose $K = 20$.

In Sect. 4 of the paper, we focus on the label/pure robustness properties with the metric $D=\mathrm{ADE}$ for simplicity. The robustness properties with $\mathrm{ADE}_K$ needs a slight modification. We state it as follows: 
\begin{thm}[Label Robustness for $\mathrm{ADE}_K$]  \label{def:labelrobustnessk}
Let $\hat {\mathbf X}=(\hat X_0,\hat X_1,\ldots,\linebreak[0] \hat X_N)$ be the past trajectories of the to-be-predicted agent and its $N$ neighbouring agents, and
$Y_\f$ its ground truth of the future trajectories of the to-be-predicted agent.
Given a prediction model $g$, an evaluation metric $D$, a safety constant $s$, then $g$ is label-robust at $\hat {\mathbf X}$ w.r.t. the perturbation radius $r > 0$ if for any $X_i\in B(\hat X_i,r)$ ($i=0,1,\ldots,N$) and any $\mathbb Y \in g(X_0,X_1,\ldots,X_N)$ with $|\mathbb Y|=K$, we have $\mathrm{ADE}_K(\mathbb Y, Y_\f) \leq s$.
\end{thm}
\begin{thm}[Pure Robustness for $\mathrm{ADE}_K$] \label{def:purerobustnessk}
Let $\hat {\mathbf X}=(\hat X_0,\hat X_1,\ldots,\linebreak[0] \hat X_N)$ be the past trajectories of the to-be-predicted agent and its $N$ neighbouring agents.
Given a prediction model $g$, an evaluation metric $D$, a safety constant $s$, then $g$ is purely robust at $\hat {\mathbf X}$ w.r.t. the perturbation radius $r > 0$ if for any $X_i\in B(\hat X_i,r)$ ($i=0,1,\ldots,N$) and any any $\mathbb Y \in g(X_0,X_1,\ldots,X_N)$ with $|\mathbb Y|=K$, there exists $\hat Y \in g(\hat {\mathbf X})$, s.t. $\mathrm{ADE}_K(\mathbb Y,\hat Y) \leq s$.
\end{thm}

The PAC guarantee constructed in Thm.~\ref{thm:main} will not be violated, where we only need to modify the measurable space as $(B(\hat{\mathbf X},r) \times \varOmega^K,\mathcal B(B(\hat{\mathbf X},r)) \times \mathcal F^K)$ for label robustness, or  $(B(\hat{\mathbf X},r) \times \varOmega^{K} \times \varOmega,\mathcal B(B(\hat{\mathbf X},r)) \times \mathcal F^{K} \times \mathcal F)$ for pure robustness. The probability $\mathbb P$, as the independent coupling, can be constructed in a quite similar way as that in Appendix B, first defined on the semi-ring of the measurable rectangles, and then uniquely extended to the $\sigma$-algebra generated by it.

\section{Experiments on SDD}
\label{appendix: Experiments on SDD}

We conduct experiments on samples from the Stanford Drone Dataset (SDD) with $r = 2$ pixels, $\eta = 0.01$ and $\epsilon = 0.01$. As shown in \cref{table: SDD}, 
\trajpac shows the similar good performance as in ETH/UCY. 
The learning time of our method in SDD is also as little as in ETH/UCY.

\begin{table}[h]
\renewcommand\arraystretch{0.95}
\scalebox{0.8}{
\begin{tabular}{ccrcccccccc}
\specialrule{.08em}{.0em}{.0em} 
\multirow{2}{*}{Scene} && \multicolumn{1}{c}{\multirow{2}{*}{ID}} && \multicolumn{3}{c}{Label Robustness}  && \multicolumn{3}{c}{Pure Robustness} \\ 
 && && Traj++ & Memo & MID  && Traj++ & Memo & MID \\ \cline{1-1} \cline{3-3} \cline{5-7} \cline{9-11}
 $quad_0$ && (84, 5) && {\color[HTML]{CB0000} \phantom{$^\dag$}\xmark$^\dag$} & {\color[HTML]{009901} \cmark} & {\color[HTML]{009901} \cmark}  && {\color[HTML]{CB0000} \phantom{$^\dag$}\xmark$^\dag$} & {\color[HTML]{009901} \cmark} & {\color[HTML]{009901} \cmark} \\
 $quad_3$ && (84, 9) && {\color[HTML]{CB0000} \phantom{$^\dag$}\xmark$^\dag$} & {\color[HTML]{F8A102} $\ocircle$} & {\color[HTML]{CB0000} \xmark}  && {\color[HTML]{CB0000} \phantom{$^\dag$}\xmark$^\dag$} & {\color[HTML]{F8A102} $\ocircle$} & {\color[HTML]{F8A102} $\ocircle$} \\
 $nexus_5$ && (588, 10) && {\color[HTML]{CB0000} \phantom{$^\dag$}\xmark$^\dag$} & {\color[HTML]{009901} \cmark} & {\color[HTML]{F8A102} $\ocircle$} && {\color[HTML]{CB0000} \phantom{$^\dag$}\xmark$^\dag$} & {\color[HTML]{F8A102} $\ocircle$} & {\color[HTML]{009901} \cmark} \\  \specialrule{.08em}{.0em}{.0em} 
\end{tabular}
}
\centering
\caption{Label/pure robustness verification on SDD with the $\mathrm{ADE}_{20}$ metric, where the safety constant is 50 pixels. Marks are the same as Tab.~\ref{table: define}.}
\label{table: SDD}
\end{table}

\section{Experiments with varying values of $r$}
\label{appendix: Experiments on r}

$r = 0.03$m is an empirical value. We chose this value as perturbation radius because it is small enough, yet it already has a significant impact on the accuracy of predictions. We also conducted experiments with $r = 0.05$ and $0.1$. Please refer to Tab.~\ref{table: radius}. As the perturbation radius grows larger, the ADE PAC bounds for different models generally expand yet they are still tight comparing to maximum sampled ADE. 
This demonstrates that the ADE PAC bound can remain unaffected with the perturbation radius increasing.

\begin{table}[h]
\centering
\scalebox{0.8}{
\begin{tabular}{ccccccccccccccc}
\specialrule{.08em}{.0em}{.0em}
methods                 && r    && label && bound & max & adver && pure && bound & max & adver \\ \cline{1-1} \cline{3-3} \cline{5-5} \cline{7-9} \cline{11-11} \cline{13-15} 
\multirow{3}{*}{Memo}   && 0.03 &&  {\color[HTML]{009901} \cmark}    && 0.98      & 0.81        & 0.73        &&  {\color[HTML]{009901} \cmark}    && 0.35      & 0.23        & 0.13        \\
                        && 0.05 &&   {\color[HTML]{F8A102} $\ocircle$}    && 1.12      & 0.87        & 0.82        &&   {\color[HTML]{009901} \cmark}   && 0.47      & 0.32        & 0.25        \\
                        && 0.1  &&   {\color[HTML]{CB0000} \xmark}    && 1.49      & 1.14        & 1.19        &&   {\color[HTML]{F8A102} $\ocircle$}   && 0.72      & 0.49        & 0.41        \\ \cline{1-1} \cline{3-3} \cline{5-5} \cline{7-9} \cline{11-11} \cline{13-15} 
\multirow{3}{*}{Traj++} && 0.03 &&  {\color[HTML]{F8A102} $\ocircle$}     && 1.01      & 0.86        & 0.65        &&   {\color[HTML]{009901} \cmark}   && 0.45      & 0.37        & 0.18        \\
                        && 0.05 &&    {\color[HTML]{F8A102} $\ocircle$}   && 1.07      & 0.86        & 0.86        &&   {\color[HTML]{009901} \cmark}   && 0.48      & 0.38        & 0.20        \\
                        && 0.1  &&  {\color[HTML]{F8A102} $\ocircle$}     && 1.12      & 0.91        & 0.84        &&   {\color[HTML]{F8A102} $\ocircle$}   && 0.57      & 0.43        & 0.14       \\  
\specialrule{.08em}{.0em}{.0em}
\end{tabular}
}
\caption{Label/pure robustness verification of Memonet/Trajectron++ on Zara1 (4430,69) (from UCY) with the corresponding ADE values (bound/max/adver indicate PAC bound/max sampled/adversarial) in three different perturbation radii.
}
\label{table: radius}
\end{table}

\end{document}